\documentclass{article}

\usepackage[english]{babel}
\usepackage{multirow}
\usepackage{amsmath,amssymb,amsfonts}
\usepackage[letterpaper,top=2cm,bottom=2cm,left=3cm,right=3cm,marginparwidth=1.75cm]{geometry}

\usepackage{amsmath}
\usepackage{graphicx}
\usepackage[colorlinks=true, allcolors=blue]{hyperref}
\DeclareMathOperator*{\argmax}{argmax} 
\DeclareMathOperator*{\argmin}{argmin}

\title{SamDSK: Combining Segment Anything Model with Domain-Specific Knowledge for Semi-Supervised Learning in Medical Image Segmentation}
\author{Yizhe~Zhang, Tao~Zhou, Shuo~Wang, Ye~Wu, Pengfei~Gu, Danny~Z.~Chen
\thanks{Y. Zhang, T. Zhou, and Y. Wu are with the School of Computer
Science and Engineering, Nanjing University of Science and Technology,
Nanjing, China. S. Wang is with the Digital Medical Research Center, School of Basic Medical Sciences, Fudan University, and Shanghai Key Laboratory of MICCAI, Shanghai, China. P. Gu and D. Chen are with the Department of Computer Science and Engineering, University of Notre Dame, Notre Dame, IN 46556, USA. \protect\\
E-mail: yizhe.zhang.cs@gmail.com, dchen@nd.edu}
}

\begin{document}
\maketitle

\begin{abstract}
The Segment Anything Model (SAM) exhibits a capability to segment a wide array of objects in natural images, serving as a versatile perceptual tool for various downstream image segmentation tasks. In contrast, medical image segmentation tasks often rely on domain-specific knowledge (DSK).
In this paper, we propose a novel method that combines the segmentation foundation model (i.e., SAM) with domain-specific knowledge for reliable utilization of unlabeled images in building a medical image segmentation model. Our new method is iterative and consists of two main stages: (1) segmentation model training; (2) expanding the labeled set by using the trained segmentation model, an unlabeled set, SAM, and domain-specific knowledge. These two stages are repeated until no more samples are added to the labeled set. A novel optimal-matching-based method is developed for combining the SAM-generated segmentation proposals and pixel-level and image-level DSK for constructing annotations of unlabeled images in the iterative stage (2). In experiments, we demonstrate the effectiveness of our proposed method for breast cancer segmentation in ultrasound images, polyp segmentation in endoscopic images, and skin lesion segmentation in dermoscopic images. Our work initiates a new direction of semi-supervised learning for medical image segmentation: the segmentation foundation model can be harnessed as a valuable tool for label-efficient segmentation learning in medical image segmentation.
\end{abstract}

\section{Introduction}
Segmentation foundational models, such as the Segment Anything Model (SAM)~\cite{kirillov2023segment}, have opened up new opportunities in medical image segmentation studies. Many recent endeavors have focused on fine-tuning and adapting SAM for medical image segmentation tasks. In cases where SAM generates high-quality segmentation proposals for medical image segmentation datasets, domain-specific knowledge can be integrated with SAM to annotate unlabeled images. This approach can be beneficial and enhance medical image segmentation, especially in scenarios where plentiful unlabeled images are accessible while manually annotated images are scarce.

\begin{figure}[h]
\centering
\includegraphics[width=0.8\textwidth]{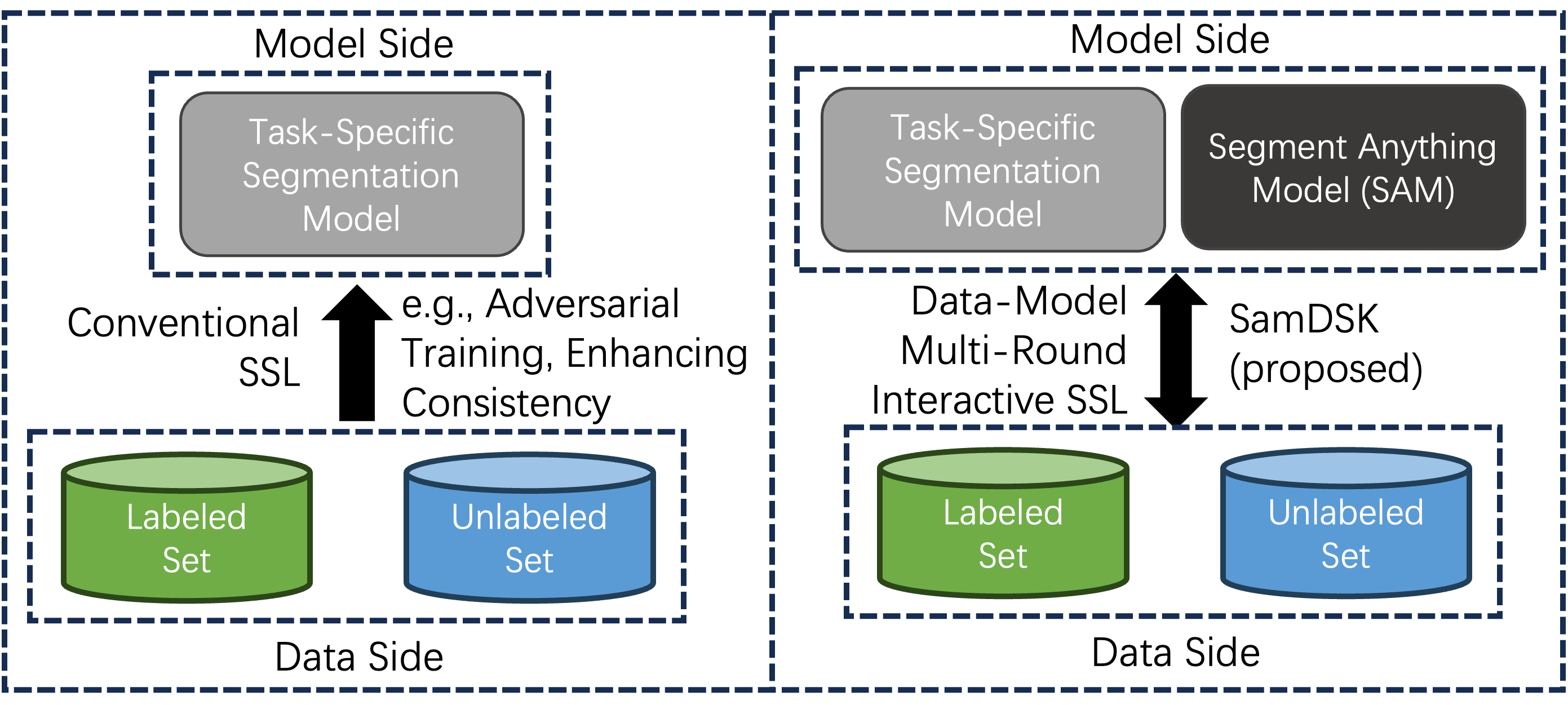}
\caption{A high-level comparison between a conventional approach (left) and our proposed SamDSK (right) for semi-supervised learning (SSL) in medical image segmentation.} \label{fig:framework}
\end{figure}

Leveraging unlabeled samples in training medical image segmentation models is a semi-supervised learning (SSL) problem, and has been extensively studied in the last decade. Zhang et al.~\cite{zhang2017deep} proposed to use adversarial training when using unlabeled samples. Adversarial training encourages a segmentation model to produce outputs for unlabeled samples that are indistinguishable from outputs for labeled samples in segmentation quality. Peiris et al.~\cite{peiris2021duo} added multi-view segmentation components to the adversarial training framework. Recently, Wang et al.~\cite{wang2023cat} proposed to incorporate anatomical constraints such as connectivity, convexity, and symmetry in adversarial training for learning from unlabeled images. Luo et al.~\cite{luo2021semi} combined a consistency criterion with adversarial training for semi-supervised learning of medical image segmentation models. Designing losses and network structures that enhance segmentation consistencies is another widely used approach for semi-supervised learning in medical image segmentation~\cite{wu2022mutual}. Another track of research for semi-supervised learning is based on pseudo-labels. Introduced in~\cite{lee2013pseudo}, pseudo-label techniques have been widely utilized for semi-supervised learning in medical image segmentation (e.g.,~\cite{rizve2020defense}). 


Adversarial training for semi-supervised learning is typically built on the assumption that the unlabeled samples are drawn from the same distribution of the labeled samples. The effectiveness of the min-max optimization in improving the segmentation quality depends on factors such as the architecture of the discriminator, the initial model parameters, the distribution changes (if any) between the unlabeled and labeled samples. Contrastive learning has shown great performance in representation learning, but, its effectiveness depends on the design of pre-tasks and the training process. Similarly, consistency loss requires identifying correct notation of consistency between unlabeled images and multiple views of the images. Incorrect or inappropriate design of the constrastive loss and consistency loss could make the learning ineffective and sometimes even harmful to the segmentation performance. Pseudo-label (PL) based techniques are often a go-to approach for utilizing unlabeled images in training a prediction model (e.g., classification model, segmentation model). Despite its effectiveness, if a large portion of the pseudo-labels is incorrect, then using PLs to train the model would suffer the risk of lower performance. 

Previous work of semi-supervised learning commonly concentrated on network training and architecture design, but fewer attempts were made on constructing and expanding the (labeled) training set. In this paper, we cast the task of utilizing unlabeled images in training a medical image segmentation model as an iterative and gradual process of annotation acquisition of unlabeled images together with model training (and re-training). Both the training set and segmentation model evolve during the learning process. We develop a novel method that combines the emerging SAM with Domain-Specific Knowledge (DSK) for annotation acquisition and labeled set expansion. More specifically, suppose a labeled image set is initially given to our method. We train an initial medical image segmentation model using the labeled images. We then apply the trained model to the unlabeled images and obtain segmentation probability maps for all the unlabeled images. We refer to these maps as pixel-level domain-specific knowledge (pixel-level DSK). We further introduce a high-level DSK called image-level DSK, which includes prior knowledge of the segmentation task, such as the potential number of RoIs (Regions of Interests) for each segmentation class in an image. For an unlabeled image, an optimization process is performed to construct its segmentation annotation by matching the segmentation proposals (generated by SAM) with the pixel-level DSK, constrained by the image-level DSK. Unlabeled images with segmentation annotations attained by the above matching process are then examined according to the optimal matching scores, and those images (with the constructed annotations) with high matching scores will be added to the labeled set as machine-labeled samples for the next round of segmentation model training. The segmentation model receives a further improvement in each round of training since more labeled data become available, and through multiple rounds of such annotation acquisition and model training/retraining, we expect more unlabeled samples with high-quality segmentation to be added to the labeled set, and thus a better medical image segmentation model will be obtained. A high-level overview of our proposed method is illustrated in Fig.~\ref{fig:framework}. 

In summary, the main contributions of this work are four-fold:
\begin{itemize}
\item[$\bullet$]  We propose a schematic shift of utilizing unlabeled images in training a medical image segmentation model. Our proposition is that once unlabeled data can be ``transformed" into labeled data, the training of the segmentation model becomes similar to the model that is trained under the fully supervised learning setup.

\item[$\bullet$] We develop a novel optimization-based method that combines the segmentation foundation model (SAM) with domain-specific knowledge (DSK) for semi-supervised learning in medical image segmentation. Our approach integrates the strengths of SAM and pseudo-label based learning, enhancing the controllability and reliability of the process of utilizing unlabeled images to train a segmentation model.

\item[$\bullet$] Given that many medical image segmentation tasks involve and benefit from image-level domain-specific knowledge, we extend our method by developing an image-level DSK within the optimization process. More specifically, we directly apply and utilize the count of regions of interest in our proposed method, aiming to regularize the process of constructing annotations for unlabeled images.

\item[$\bullet$] Analyses and experiments confirm the effectiveness of our proposed method. We also illustrate and discuss current limitations and future research directions for utilizing SAM to leverage unlabeled images in medical image segmentation.
\end{itemize}





\begin{figure*}[t]
\centering
\includegraphics[width=0.8\textwidth]{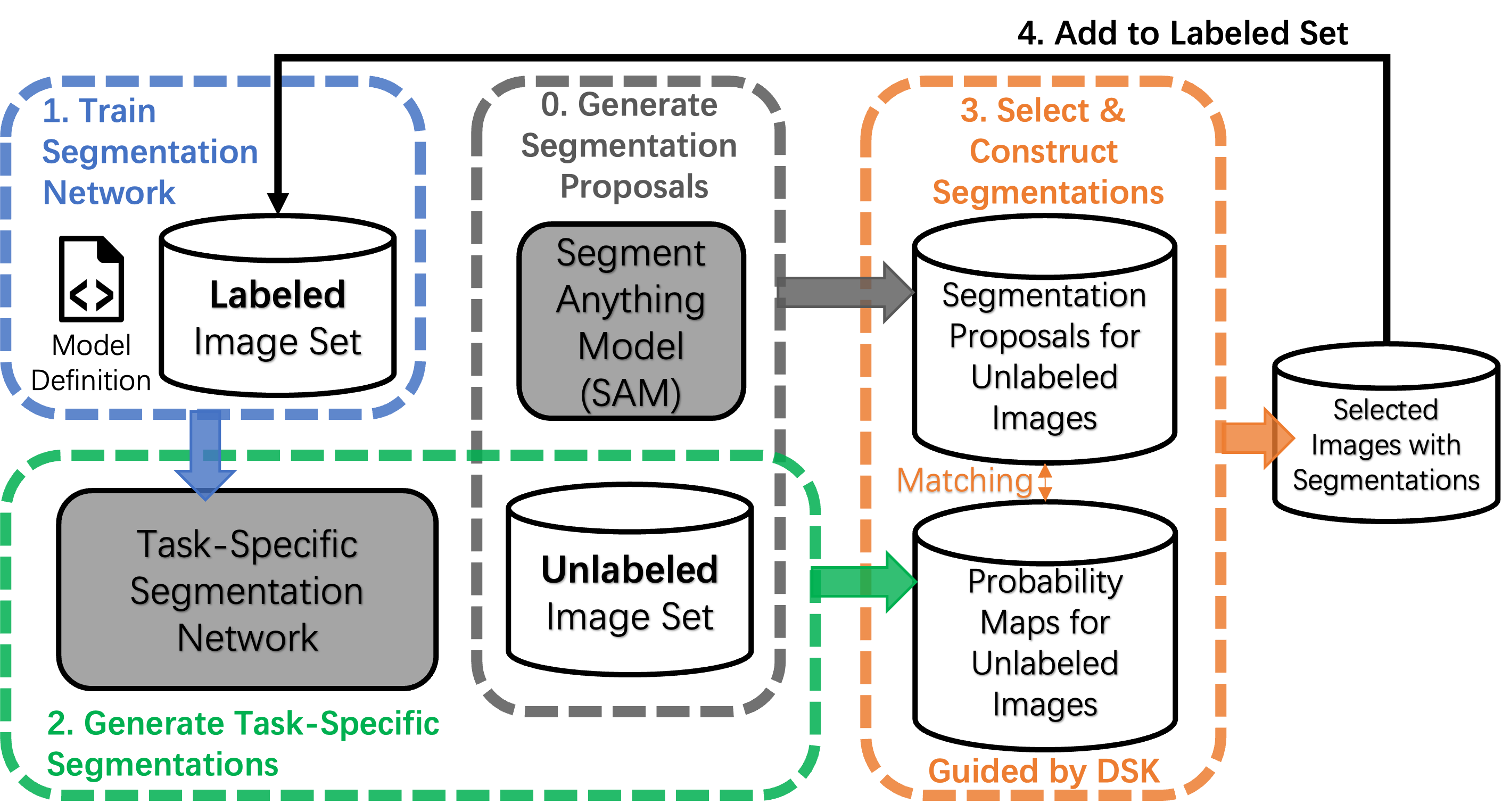}
\caption{The main steps of our proposed SamDSK method. Step-0 to step-4 are repeated iteratively until a certain criterion is met (e.g., no more samples are added to the labeled set). Step-0 may need to run only once if SAM remains unchanged during the process.} \label{fig:overview}
\end{figure*}

\section{Related Work}

\subsection{Pseudo-labels}
Our proposed SamDSK uses probability maps generated by a medical image segmentation network for unlabeled images. Although probability maps are not exactly taken as labels or pseudo-labels, these maps are often referred to as ``soft'' pseudo-labels. Hence, below we give a review of the methods that use pseudo-labels for semi-supervised learning in medical image segmentation.

Wu et al.~\cite{wu2022mutual} proposed Mutual Consistency Learning  (MCL) for utilizing unlabeled samples in segmentation model training. MCL creates multiple decoders which all share one encoder for generating segmentations. Because of the random initialization, these decoders generate possibly different segmentation predictions, and the differences between these predictions are used to compute prediction uncertainties. In addition,  mutual learning~\cite{zhang2018deep} is performed during model training to enforce the output from each decoder to be close (consistent) to the pseudo-labels generated by the other decoders. At the end of the training, only one decoder is used for model deployment. Since pseudo-labels inherently contain errors, Thompson et al.~\cite{thompson2022pseudo} proposed to use super-pixels to refine pseudo-labels in semi-supervised learning. The proposed process largely follows the classical approach of using pseudo-labels for SSL~\cite{lee2013pseudo}, with additional steps which refine pseudo-labels according to super-pixels. The main motivation of this method is to impose a spatial structure induced from the super-pixels for regularizing and refining pseudo-labels so that pseudo-labels would yield better quality and benefit the subsequent segmentation model training. Seibold et al.~\cite{seibold2022reference} proposed to use reference images (with annotations) for generating pseudo-labels. This approach works for medical image segmentation tasks with well-defined and stable object structures (e.g., chest radiographic anatomy segmentation). For tasks under more dynamic scenes where objects can be at arbitrary locations and with large variations of shapes (e.g., polyp segmentation in endoscopic images), reference-based pseudo-label generation could become less reliable and effective. Recently, Basak et al.~\cite{basak2023pseudo} proposed to utilize pseudo-labels in conjunction with contrastive learning. 

Our proposed method utilizes probability maps generated by a task-specific medical image segmentation model. Similar to~\cite{thompson2022pseudo}, we employ an external component (i.e., SAM in our case) to work with the medical image segmentation model. Instead of refining pseudo-labels using super-pixels, we aim to use DSK to match and select SAM-generated segmentation proposals and construct segmentations for unlabeled images for model re-training. Comparing to the super-pixel based methods, SAM is a much more advanced model and using SAM allows us to design a more controllable optimization process (e.g, multiple rounds of optimal matching) in the usage of domain-specific knowledge both at the pixel level and image level for generating segmentations of unlabeled images.

\subsection{SAM for Medical Image Segmentation}
Since the introduction of the Segment Anything Model (SAM), there has been a wave of attempts to develop a more accurate and robust medical image segmentation system with the utilization of SAM~\cite{huang2023segment,zhou2023can,ma2023segment,wu2023medical,zhang2023survey,qiao2023robustness}. Recent work has shown that SAM alone, without further fine-tuning and/or adaptation, often delivers unsatisfactory
 results for medical image segmentation tasks \cite{huang2023segment,zhou2023can}. To utilize SAM more effectively, Ma et al.~\cite{ma2023segment} proposed to fine-tune SAM using labeled images. Wu et al.~\cite{wu2023medical} proposed to add additional layers to adapt SAM for a medical image segmentation task. Our work aims at using SAM for semi-supervised learning of a medical image segmentation model.

\subsection{SAM and Pseudo-labels}
Recently, Chen et al.~\cite{chen2023segment} proposed to use SAM to refine pseudo-labels in weakly-supervised learning. SAM is employed in three places: (1) refining CAM-generated attention maps, (2) refining segmentation generated by post-processing, and (3) refining segmentations generated by a segmentation model (DeepLab). In this paper, we utilize probability maps as pixel-level DSK for selecting segmentation proposals produced by SAM. To make the selection more reliable, we further employ an image-level DSK (e.g., the potential number of regions of interests) during the selection process to constrain the optimization process as a way of realizing the idea of combining SAM with domain-specific knowledge (at both the pixel level and image level). 

\begin{figure*}[t]
\centering
\includegraphics[width=0.9\textwidth]{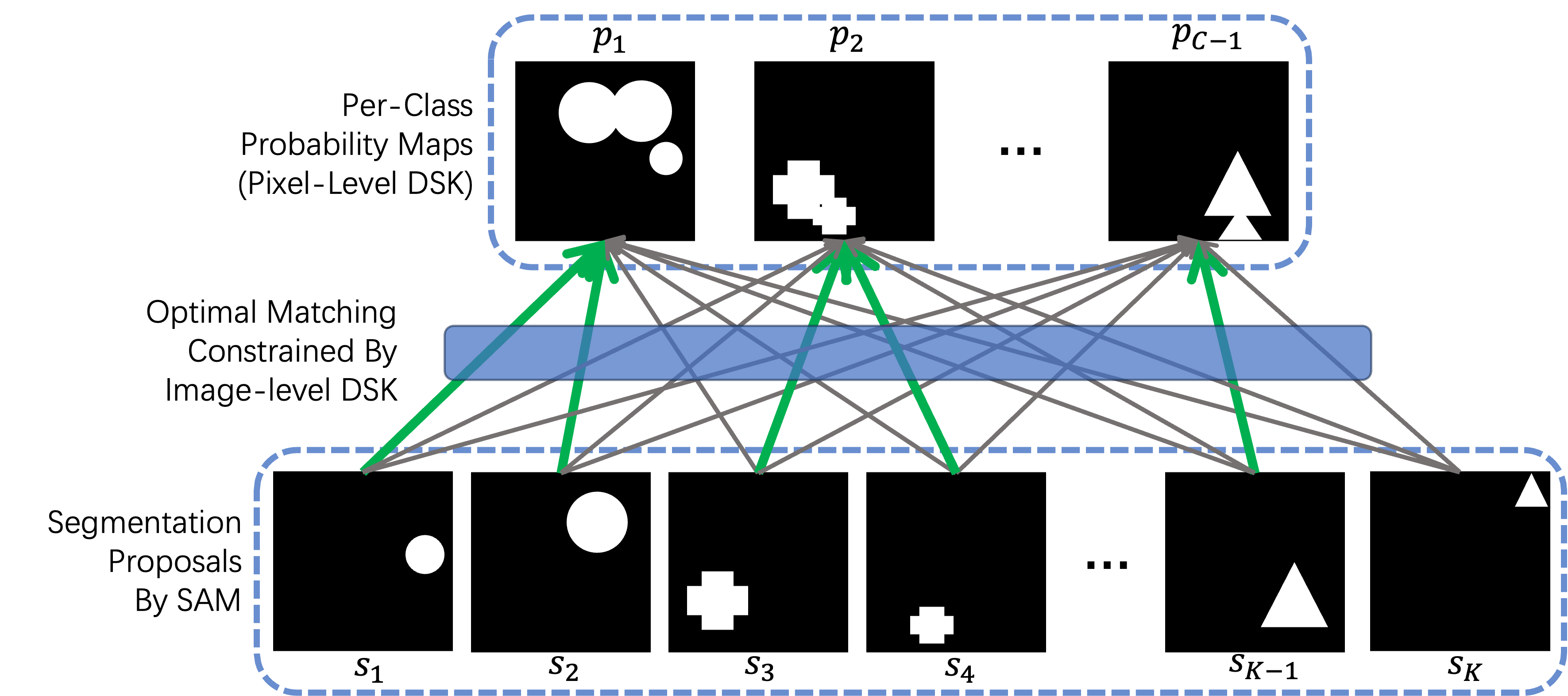}
\caption{Matching segmentation proposals generated by SAM with pixel-level DSK generated by the current trained segmentation model. The matching is constrained by additional image-level DSK, i.e., potential counts of RoIs. Green arrows mark the obtained solution.}\label{fig:matching}
\end{figure*}

\section{Methodology}
The input to our SamDSK model consists of four components: a labeled image set $D = \{(x_1,y_1), (x_2,y_2), \dots, (x_n, y_n)\}$, an unlabeled image set $U=\{u_1, u_2, \dots, u_m\}$, Segment Anything Model $\textrm{SAM}$, and a medical image segmentation model ${M}$ with an encoder initialized using ImageNet~\cite{imagenet} pre-trained weights and a decoder initialized randomly. SamDSK performs five main steps (step-0 to step-4). An overview of the five steps and their relations are illustrated in Fig.~\ref{fig:overview}. In Sections~\ref{sec:m2}, \ref{sec:m3}, and \ref{sec:m4}, we describe the key step (step-3) for generating annotations of unlabeled images. In Section~\ref{sec:m5}, we present the process of segmentation model training using both human-labeled and machine-labeled images. In Section~\ref{sec:m6}, we analyze the dynamics of our proposed multi-round SSL method. Finally, in Section~\ref{sec:m7}, we give discussions of several medical image segmentation tasks that SamDSK is applicable to.

\subsection{Applying SAM to Unlabeled Medical Images}
Given a set $U$ of unlabeled medical images, we apply the state-of-the-art SAM to these images to obtain segmentation proposals in each image. A key consideration in this step is to use relatively lower threshold settings in SAM to ensure the inclusion of all potential Regions of Interest (RoIs) in the generated segmentation proposals. By default, we use the following parameter values when employing SAM: ``crop\_nms\_threshold = 0.5, box\_nms\_thresh = 0.5, pred\_iou\_thresh = 0.5, stability\_score\_thresh = 0.5". With the ongoing development of the segmentation foundation model, future SAM variants are expected to provide improved coverage of RoIs for various medical image segmentation tasks.

\subsection{Matching Segmentation Proposals with Pixel-level DSK}\label{sec:m2}
Suppose we already train a segmentation model ${M}$ using the currently available labeled data. Also, suppose for an unlabeled image $u$ with width $w$ and height $h$, SAM provides a set of $K$ segmentation proposals, $S=\{s_1, s_2, \dots, s_K\}$, where $s_i \in \{0,1\}^{h\times w}$ for $i=1, 2, \dots, K$. The segmentation model ${M}$ gives $C$ distinct probability maps $\{p_1, p_2, \dots, p_C\}$, where $p_c \in \mathbb{R}^{h\times w}$ for $c = 1, 2, \dots, C$. $C$ is the number of classes in the segmentation task. The background class is taken always as the class $C$.
For each class $c\in\{1,2, \dots, C\}$, we create a set of binary scalar indicators, $Z_c=\{z_{c,1}, z_{c,2}, \dots, z_{c,K}\}$, where $z_{c,k} \in \{0,1\}$ for $k=1, 2, \dots, K$. Each indicator $z_{c,k}$ is multiplied with the corresponding segmentation proposal $s_k$ in constructing the segmentation map for class $c$. We aim to find an optimal configuration of $z_{1,1}, z_{1,2}, \dots, z_{C,K}$ which maximizes the total sum of IoU scores between the segmentation proposals and the segmentation probability maps, as:
\begin{equation} \label{eq1}
z_{1,1}^*, \dots, z_{C,K}^* = \argmax_{z_{1,1}, z_{1,2}, \dots, z_{C,K}} \sum_{c=1}^{C-1} \textrm{IoU} (p_c, \tau(\sum_{k=1}^{K}z_{c,k}s_k)),
\end{equation}
\begin{equation} \label{eq2}
\begin{aligned} 
s.t., &\sum_{c=1}^{C-1} z_{c,k} \leq 1,\\
& z_{c,k} \in \{0,1\}, \ c=1,2, \dots, C, \ k=1,2, \dots, K,
\end{aligned}
\end{equation}

\noindent
where $\tau(\cdot)$ converts all values which are larger than 1 to 1, and $\textrm{IoU}(\cdot,\cdot)$ computes the Intersection over Union score between a probability map and a segmentation proposal. The constraints in Eq.~(\ref{eq2}) ensure that each segmentation proposal can be assigned to only one class label. Note that there can be segmentation proposals not being selected in forming the optimal matching. 

In Fig.~\ref{fig:matching}, we provide a high-level illustration of the mechanism that we use for this optimal matching formulation. The optimization problem in Eq.~(\ref{eq1}) (with the constraints in Eq.~(\ref{eq2})) is a maximum bipartite matching problem which can be efficiently solved in polynomial time~\cite{Bang-Jensen2000}. 


\subsection{Incorporating Image-level DSK}\label{sec:m3}
In this section, we aim to improve the effectiveness and reliability of the above proposed segmentation annotation construction by adding image-level DSK as new constraints to the optimal matching formulation. Suppose the number of RoIs for an image of a given medical image segmentation task ranges from $v_c^{lower}$ to $v_c^{upper}$ for class $c$. We add these constraints to Eq.~(\ref{eq2}). The optimization objective remains unchanged, but the  constraints are updated with additional constraints imposed by $v_c^{lower}$ and $v_c^{upper}$. Formally, we aim to optimize the objective in Eq.~(\ref{eq1}), with the following constraints:
\begin{equation}\label{eq3}
\begin{aligned}  
s.t., &\sum_{c=1}^{C-1} z_{c,k}\leq1,\\ 
&v_c^{lower} \leq \sum_{k=1}^K z_{c,k}\leq v_c^{upper}, c \in\{1,2, \dots, C-1\}, \\
& z_{c,k} \in \{0,1\}, \ c=1,2, \dots, C, \ k=1,2, \dots, K.
\end{aligned} 
\end{equation}

In each round of the model training and labeled set expansion, we control the $v_c^{lower}$ and $v_c^{upper}$ values to ensure a stable way of gradually including more samples with more RoIs into the labeled set. For the first round, we can set $v^{lower}$ and $v^{upper}$ as both equal to 1. This indicates that the optimal matching is only a one-to-one matching between the segmentation proposals and the probability maps. In the second round, we can set $v_c^{lower}$ as 1 and $v_c^{upper}$ as 2 to allow a two-to-one matching. Depending on the specific segmentation task, we can use a larger increment of $v_c^{upper}$ to allow a more effective inclusion of unlabeled samples into the labeled set.

\begin{figure*}[t]
\centering
\includegraphics[width=0.9\textwidth]{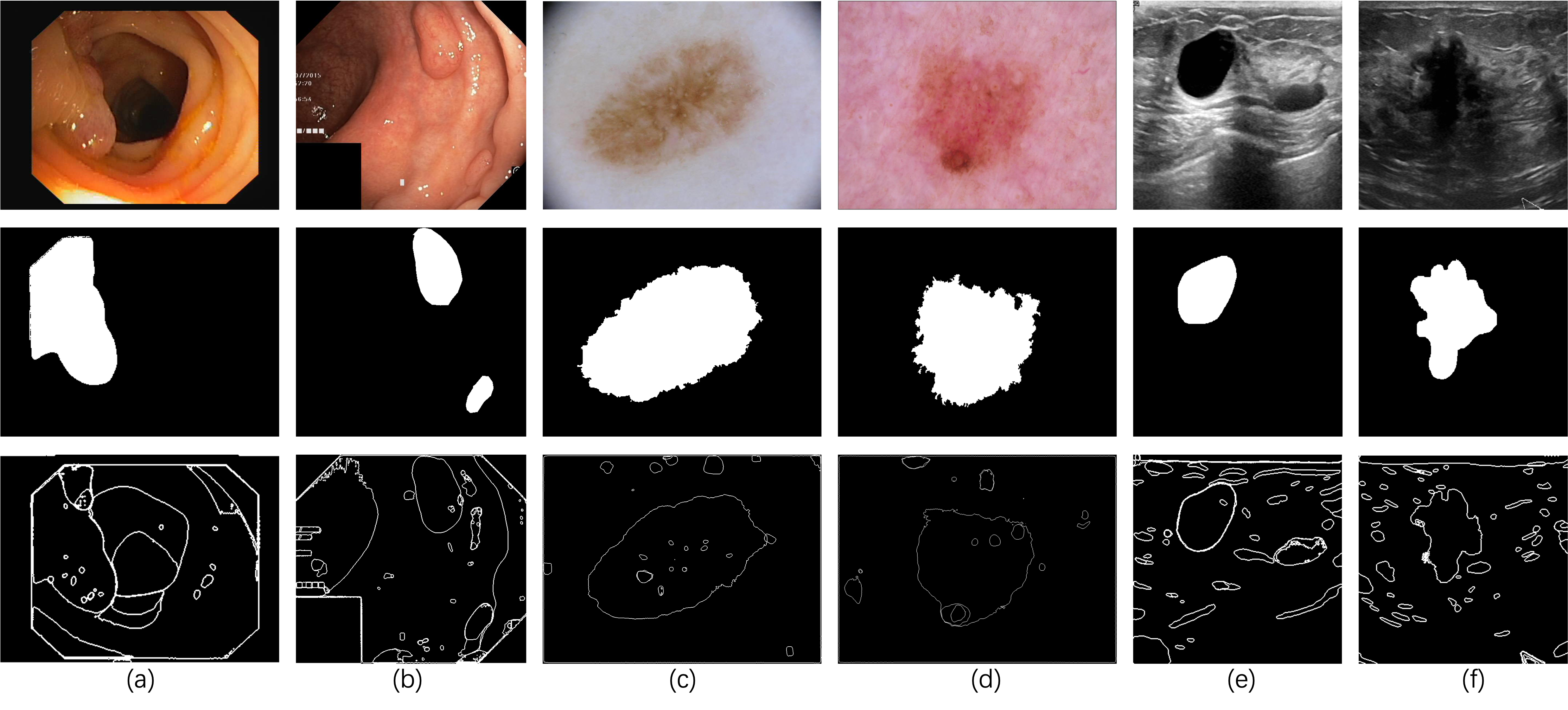}
\caption{Image examples of polyp segmentation ((a) and (b)), skin lesion segmentation ((c) and (d)), and breast cancer segmentation ((e) and (f)). Row 1: raw images. Row 2: ground truth annotations. Row 3: Boundary maps of the SAM-generated segmentation proposals.} \label{fig:example_overview}
\end{figure*}

\subsection{Segmentation Annotation Construction}\label{sec:m4}
With the identified values $z_{1,1}^*, z_{1,2}^*, \dots, z_{C,K}^*$ through optimizing Eq.~(\ref{eq1}) and Eq.~(\ref{eq3}), we compute the overall score $\beta$ for the image, as follows: 
\begin{equation}\label{eq:beta}
\begin{aligned}  
\beta = \frac{1}{C-1}\sum_{c=1}^{C-1} \textrm{IoU} (p_c, \tau(\sum_{k=1}^{K}z_{c,k}^*s_k)).
\end{aligned} 
\end{equation}

A sample with a $\beta$ score higher than a pre-defined threshold (denoted as $\beta^*$; its default value is 0.9) is selected to advance to the next round of segmentation model training. For a selected sample, its annotation map is constructed as $q_c=\tau(\sum_{k=1}^{K}z_{c,k}^*s_k)$, for $c = 1, 2, \dots, C-1$. Pixel areas not covered by any segmentation proposals selected by the optimal matching are considered as belonging to the background class (class $C$).



\subsection{Segmentation Model Training}\label{sec:m5}
Suppose the unlabeled samples with their annotation maps generated by the above optimization procedure are denoted as ${D^{machine}} = \{(u_j, q_{j,c})\}$, for $j = 1, 2, \dots, m'$, $m'\leq m$, and $c= 1, 2, \dots, C$. The original human-labeled set is denoted as $D = \{(x_i, y_{i,c})\}$, for $i = 1, 2, \dots, n$, $c= 1, 2, \dots, C$. We aim to optimize the following objective function with respect to the parameters $\theta$ of a medical image segmentation model ${M}$, as: 
\begin{equation}
\argmin_{\theta}(\sum_{c=1}^{C}\sum_{i=1}^{n}{L}(M(x_i,y_{i,c})) + \lambda \sum_{c=1}^{C}\sum_{j=1}^{m'}{L}(M(u_j,q_{j,c}))),
\end{equation}
where $\lambda$ is set as 1 by default. Assuming that the medical image segmentation model ${M}$ generates probability maps for each class, the loss function $L$ can take the form of a region-based loss (e.g., Dice loss), a pixel-level entropy-based loss, or a combination of these two types of losses. A mini-batch based stochastic gradient descent method (e.g., Adam) can be applied to optimize the loss via updating the parameters $\theta$ of the 
model ${M}$.

\subsection{Analyses}\label{sec:m6}
The SamDSK method combines SAM-generated segmentation proposals with domain-specific knowledge both at the pixel-level and image-level in order to generate annotations for unlabeled images. Pixel-level DSK is provided by the segmentation model ${M}$ trained on the current labeled image set, and optimal matching is performed with constraints imposed by image-level DSK as well as the base constraints that only one segmentation proposal can be assigned to one class label. There exist clearly two cases after the optimal matching process for constructing annotations of the unlabeled images.

\begin{itemize}
\item Case-1: Prediction maps produced by the current segmentation model ${M}$ match well with a subset of the segmentation proposals (generated by SAM). In this scenario, the optimal matching gives a high $\beta$ matching score (i.e., higher than $\beta^*$), and this image with the constructed annotations is added to the labeled image set for the subsequent rounds of segmentation model retraining.

\item Case-2: Prediction maps produced by the current segmentation model ${M}$ do not match well with any subsets of the segmentation proposals generated by SAM. In this situation, optimal matching gives a low $\beta$ matching score, and this image is not added to the labeled set for the next round of model retraining. 

\end{itemize}

For simplicity, we use a binary segmentation task to illustrate, and focus on its segmentation class \#1 (the foreground class) to provide the analyses below. The same logic applies to multi-class segmentation tasks with multiple classes of foreground objects.

\textbf{Assumption 1:} For an unlabeled image $u$, there exists a subset of segmentation proposals, $\Gamma_1$, generated by SAM, such that the union of its elements closely approximates the ground truth of segmentation class 1 in $u$.\footnote{SamDSK is not yet ready to be applied to those medical image segmentation tasks for which SAM fails to generate sufficiently good segmentation proposals.} More formally, this assumption can be described as: 
\begin{equation}
\textrm{IoU}(\bigcup_{s \in \Gamma_1} s, gt_1) > 1-\epsilon,
\end{equation}
where $gt_1$ represents the ground truth annotation map of class 1 for the image $u$, and $\epsilon$ is a small positive value (e.g., $\epsilon$ = 0.02). It is important to note that we do not have access to the ground truth annotations of the unlabeled images.

\vspace{0.06in}
\noindent
\textbf{Proposition 1:} If an unlabeled image falls into the Case-2 category described above, then its probability maps generated by the segmentation model ${M}$ are not closely aligned with their corresponding ground truth annotation maps.

\vspace{0.06in}
\noindent
{\bf Proof:} For an unlabeled image in Case-2, the optimal matching identifies no subset of the segmentation proposals that yields a sufficiently high Intersection over Union (IoU) score (e.g., a score higher than 0.9) for the probability map of class 1 (i.e., $p_1$) generated by the segmentation model ${M}$. Given that $\Gamma_1$ is a feasible solution attained when seeking an optimal matching, it follows that the union of elements in $\Gamma_1$ is not well-matched with the probability map $p_1$. This can be formally expressed as:
\begin{equation}
\textrm{IoU}(\bigcup_{s \in \Gamma_1} s, p_1) < \beta^*.
\end{equation}

Since the union of elements in $\Gamma_1$ closely approximates the ground truth and $p_1$ is not well aligned with the union of elements in $\Gamma_1$, it follows that $p_1$ is NOT closely aligned with the ground truth annotation map $gt_1$. Formally, an upper bound of the Intersection over Union (IoU) between the ground truth annotation map and the prediction map of class \#1 is given by:
\begin{equation}
\textrm{IoU}(gt_1, p_1) \lesssim \beta^* + \epsilon.
\end{equation}

\hfill 
\vspace{0.06in}

With the above argument, we have demonstrated that the unlabeled samples which are not added to the labeled set (Case-2) do not yet contain sufficiently accurate annotations (e.g., $p_1$ is not sufficiently accurate with respect to the ground truth). On the other hand, for the unlabeled images with annotations that are added to the labeled set (Case-1), it is still possible that although their annotations match well with a subset of the segmentation proposals, denoted as $\Gamma_1$, the ground truth annotation map is actually closer to another subset $\Gamma_1'$ of the segmentation proposals, and $\Gamma_1$ may or may not be equal to $\Gamma_1'$. Consequently, we cannot provide an explicit guarantee on the correctness of $p_1$ and its corresponding $\bigcup_{s \in \Gamma_1} s$ for the samples in Case-1. Nevertheless, in our experiments, we present empirical evidence to demonstrate that the annotations of samples in Case-1 (those added to the labeled set) are closer to their corresponding ground truth annotations than those in Case-2.



\begin{table}[!]
\centering
\caption{Breast cancer segmentation in ultrasound images (in Dice coefficient). The best results are marked in {\bf bold}, and the second best results are \underline{underlined}. The red color highlights performance degradation compared to the baseline. The same markings are used in the other tables.}
\begin{tabular}{|c|c|c|c|c|}
\hline
 \textbf{Method} & Network & Labeled & Unlabeled & Dice   \\\hline
Baseline & \multirow{6}{*}{TransUNet} & \multirow{6}{*}{30\%} & 0\% & 52.7\\
\cline{1-1}
\cline{4-5}
PL~\cite{lee2013pseudo} &  & &  \multirow{5}{*}{70\%} & \textcolor{red}{50.8} \\
\cline{1-1}
\cline{5-5}
SP-Refine~\cite{thompson2022pseudo} &  & & & 53.2 \\
\cline{1-1}
\cline{5-5}
BCP~\cite{bai2023bidirectional} &  & &  & 53.7 \\
\cline{1-1}
\cline{5-5}
UPS~\cite{rizve2020defense} &  & &  & \underline{54.3} \\
\cline{1-1}
\cline{5-5}
 SamDSK (ours) &  &  & & \textbf{59.4}\\
\hline
\hline
Baseline & \multirow{6}{*}{HSNet} & \multirow{6}{*}{30\%} & 0\% & 64.3 \\
\cline{1-1}
\cline{4-5}
PL~\cite{lee2013pseudo} &  & & \multirow{5}{*}{70\%} & 69.5 \\
\cline{1-1}
\cline{5-5}
SP-Refine~\cite{thompson2022pseudo} &  & & & 69.7 \\
\cline{1-1}
\cline{5-5}
BCP~\cite{bai2023bidirectional} &  & & & \textcolor{red}{64.1} \\
\cline{1-1}
\cline{5-5}
UPS~\cite{rizve2020defense} &  & & & \underline{70.8} \\
\cline{1-1}
\cline{5-5}
 SamDSK (ours) &  & & & \textbf{73.6}\\
 \hline

\end{tabular}
\label{BCS_results}
\end{table}

\begin{table*}[!]
\centering
\footnotesize
\caption{Segmentation performances of different methods on five polyp segmentation datasets (in Dice coefficient).}
\begin{tabular}{|c|c|c|c| c | c | c| c| c|}
\hline
 \textbf{Method} & Network & Labeled & Unlabeled & CVC-300 & CVC-ClinicDB & Kvasir & CVC-ColonDB & ETIS  \\\hline
Baseline & \multirow{5}{*}{PraNet} & \multirow{5}{*}{10\%} & 0\% & {82.3} & 76.2 & 83.1 & 63.7 &{60.2}\\
\cline{1-1}
\cline{4-9}
PL~\cite{lee2013pseudo} &  &  & \multirow{4}{*}{90\%} &  {83.8} & {77.7} & 85.1 & \textcolor{red}{63.4} & {63.3}\\
\cline{1-1}
\cline{5-9}
BCP~\cite{bai2023bidirectional} & & & & 83.4 & {78.3} & {85.3} & \underline{65.6} &\textbf{64.3}\\
\cline{1-1}
\cline{5-9}
UPS~\cite{rizve2020defense} & & & & \underline{84.3} & \underline{81.7} & \underline{86.6} & \textcolor{red}{62.3} &{58.2}\\
\cline{1-1}
\cline{5-9}
 SamDSK (ours) &  & & & \textbf{89.4} & \textbf{84.4} & \textbf{88.0} & \textbf{67.6} & \underline{63.5}\\
\hline
 \hline
Baseline & \multirow{5}{*}{HSNet} & \multirow{5}{*}{10\%} & 0\% & {85.1} & 81.7 & 86.9 & 66.4 &{68.9}\\
\cline{1-1}
\cline{4-9}
PL~\cite{lee2013pseudo} &  &  & \multirow{4}{*}{90\%} & {87.5} & {84.3} & 89.6 & 71.5 & \underline{72.9}\\
\cline{1-1}
\cline{5-9}
BCP~\cite{bai2023bidirectional} &  & & & {85.8} & 84.9 & 88.1 & 68.5 & \textcolor{red}{68.3}\\
\cline{1-1}
\cline{5-9}
 UPS~\cite{rizve2020defense} &  & & & \textbf{88.3} & \textbf{86.3} & \underline{89.8} & \underline{72.9} & \textbf{76.7}\\
\cline{1-1}
\cline{5-9}
 SamDSK (ours) &  & & & \textbf{88.3} & \underline{85.2} & \textbf{90.2} & \textbf{76.6} & {72.7}\\
 \hline
 \hline
Baseline & \multirow{5}{*}{PraNet} & \multirow{5}{*}{30\%} & 0\% & {89.7} & 83.2 & 86.5 & 63.2 & \underline{64.7}\\
\cline{1-1}
\cline{4-9}
PL~\cite{lee2013pseudo} &  &  & \multirow{4}{*}{70\%} & \underline{90.8} & \textbf{86.7} & \underline{88.3} & 68.5 & \textcolor{red}{61.1}\\
\cline{1-1}
\cline{5-9}
BCP~\cite{bai2023bidirectional} &  & & & {\textcolor{red}{88.3}} & 84.3 & \textbf{89.1} & \underline{69.0} & \textbf{69.7}\\
\cline{1-1}
\cline{5-9}
UPS~\cite{rizve2020defense} &  & & & {\textcolor{red}{87.8}} & 84.2 & {87.1} & 64.8 & \textcolor{red}{58.3}\\
\cline{1-1}
\cline{5-9}

 SamDSK (ours) &  & & & \textbf{91.7} & \underline{85.2} & {87.6} & \textbf{71.4} & \underline{67.4}\\
\hline
 \hline
Baseline & \multirow{5}{*}{HSNet} & \multirow{5}{*}{30\%} & 0\%&{87.1} & 86.5 & 90.9 & 74.9 &{76.4}\\
\cline{1-1}
\cline{4-9}
PL~\cite{lee2013pseudo} & & & \multirow{4}{*}{70\%} & 87.8 & {88.9} & {91.2} & {77.3} &76.9\\
\cline{1-1}
\cline{5-9}
BCP~\cite{bai2023bidirectional} & & &  & \underline{88.3} & 88.4 & 90.9 & 75.5 & \textcolor{red}{74.0}\\
\cline{1-1}
\cline{5-9}
UPS~\cite{rizve2020defense} & & &  & {87.6} & \underline{89.4} & \underline{91.8} & \underline{77.8} & \textbf{78.5}\\
\cline{1-1}
\cline{5-9}
 SamDSK (ours) &  & &  & \textbf{88.9} & \textbf{89.6} & \textbf{92.0} & \textbf{79.8} & \underline{77.5}\\
 \hline

\end{tabular}
\label{polyp_results}
\end{table*}

\begin{table}[!]
\centering
\caption{Skin lesion segmentation on the ISIC 2018 dataset (in Dice coefficient).}
\begin{tabular}{|c|c|c|c|c|}
\hline
 \textbf{Method} & Network & Labeled & Unlabeled & Dice   \\\hline
Baseline & \multirow{5}{*}{PraNet} & \multirow{5}{*}{10\%} & 0\% & 86.1 \\
\cline{1-1}
\cline{4-5}
PL~\cite{lee2013pseudo} &  &  & \multirow{4}{*}{90\%} & 87.3 \\
\cline{1-1}
\cline{5-5}
SP-Refine~\cite{thompson2022pseudo} &  &  &  & 87.2 \\
\cline{1-1}
\cline{5-5}
BCP~\cite{bai2023bidirectional} &  & &  & \underline{87.5} \\
\cline{1-1}
\cline{5-5}
 SamDSK (ours) &  &  & & \textbf{88.1}\\
\hline
\hline
Baseline & \multirow{5}{*}{HSNet} & \multirow{5}{*}{10\%} & 0\% & 86.5 \\
\cline{1-1}
\cline{4-5}
PL~\cite{lee2013pseudo} &  &  & \multirow{4}{*}{90\%} & 87.7 \\
\cline{1-1}
\cline{5-5}
SP-Refine~\cite{thompson2022pseudo} &  &  & &  88.2 \\
\cline{1-1}
\cline{5-5}
BCP~\cite{bai2023bidirectional} &  &  & & \underline{88.4} \\
\cline{1-1}
\cline{5-5}
 SamDSK (ours) &  &  & & \textbf{89.9}\\
\hline

\end{tabular}
\label{isic_results}
\end{table}

\begin{figure}[t]
\centering
\includegraphics[width=0.45\textwidth]{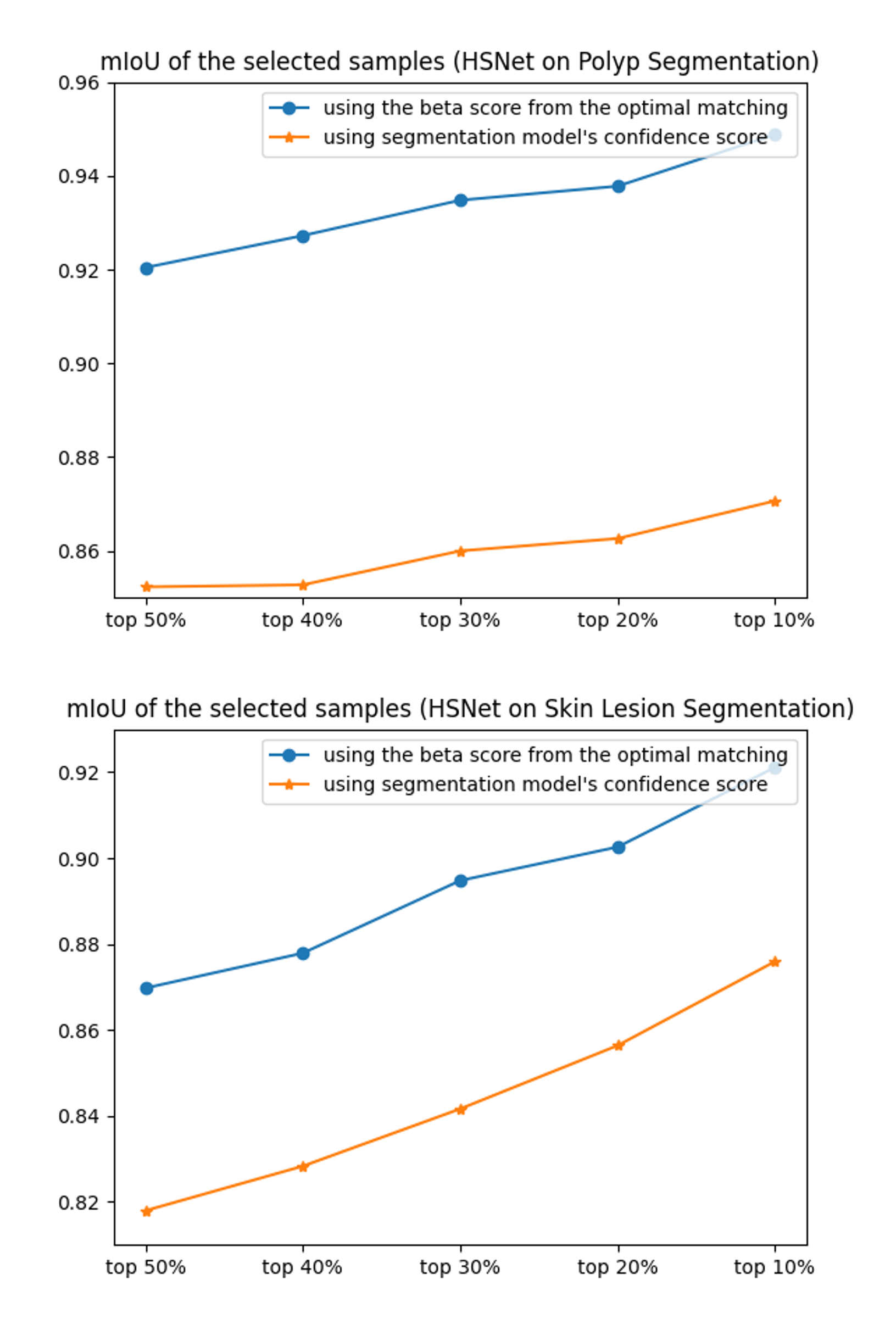}
\caption{Mean IoU (mIoU) of segmentation predictions and ground truth annotations for selected samples. The $\beta$ score (in Eq.~(\ref{eq:beta})) enables SamDSK to select samples with higher segmentation quality (higher mIoU) compared to samples selected by using the model's own uncertainty and confidence measures (as conducted in UPS~\cite{rizve2020defense}).} \label{fig:sample2}
\end{figure}

\begin{figure}[t]
\centering
\includegraphics[width=0.55\textwidth]{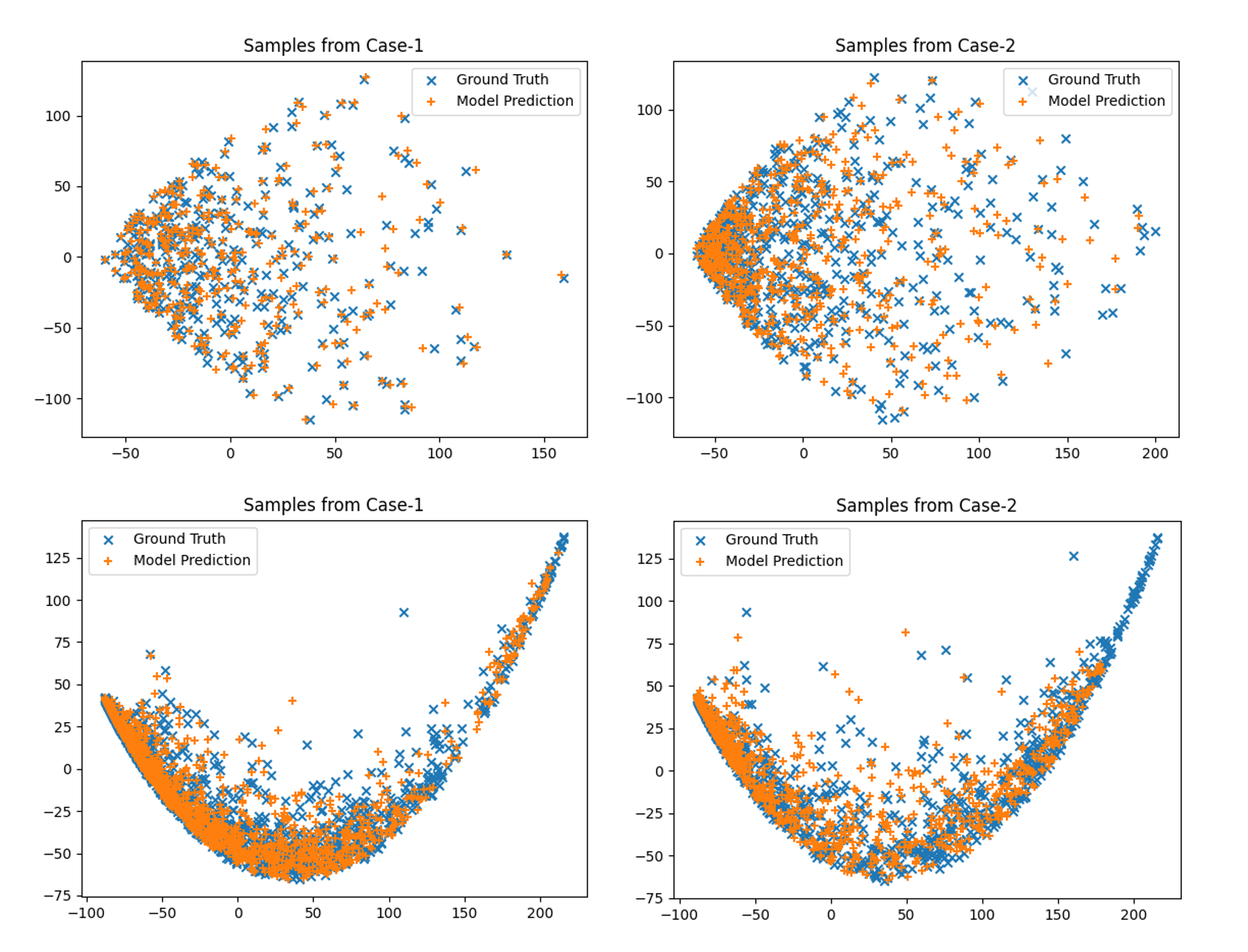}
\caption{Ground truth annotations and segmentation predictions after dimensionality reduction (PCA). The segmentation predictions for samples in Case-1 are closer to the ground truth annotations than those in Case-2. Top row: samples of the polyp segmentation dataset. Bottom row: samples of the skin lesion segmentation dataset. } \label{fig:points}
\end{figure}

\begin{figure}[t]
\centering
\includegraphics[width=0.39\textwidth]{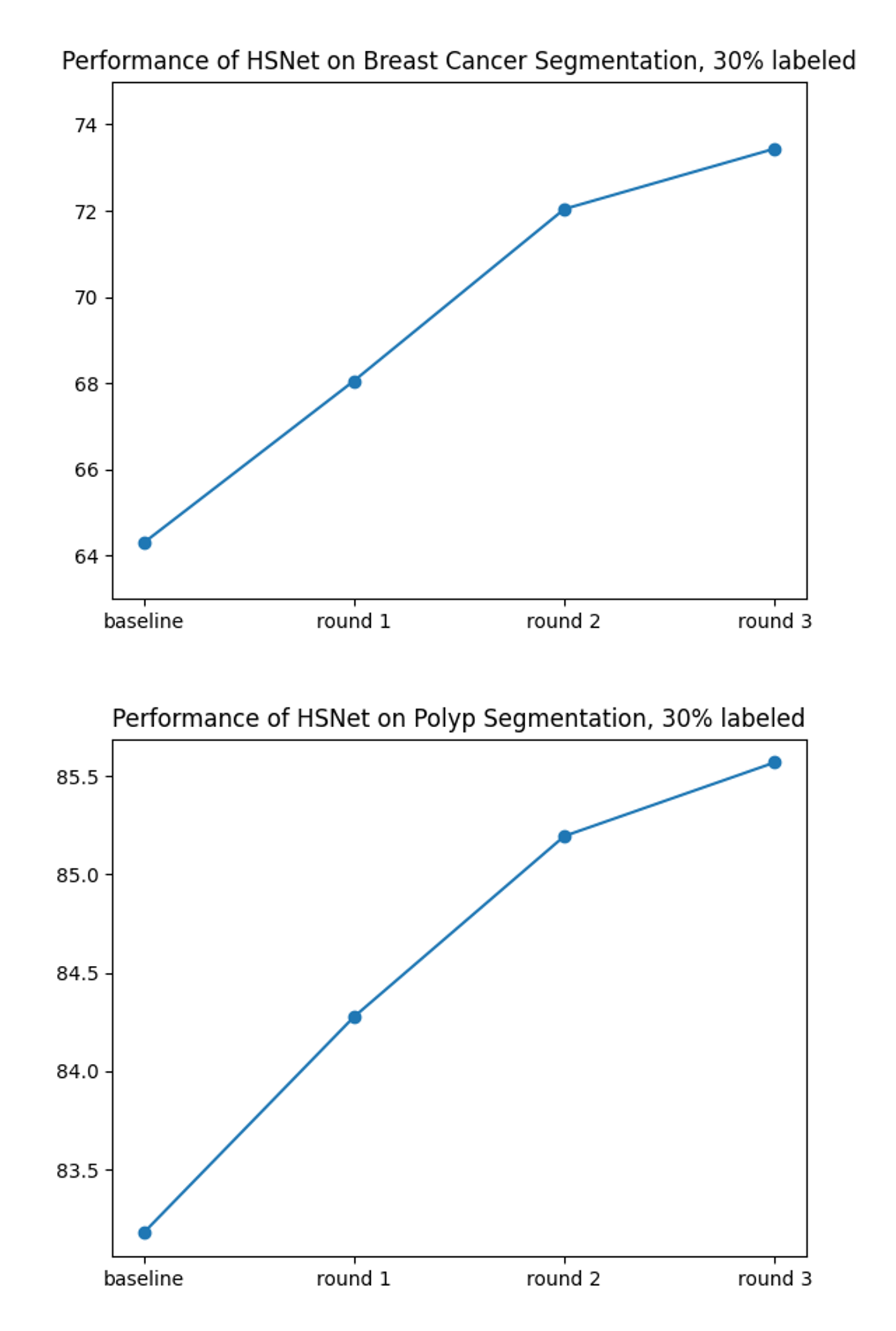}
\caption{Segmentation performances (in Dice coefficient) across different rounds of processing in SamDSK.} \label{fig:SamDSK_plot1}
\end{figure}

\subsection{Applicable Segmentation Tasks}\label{sec:m7}
Medical image segmentation encompasses a multitude of imaging modalities (e.g., MRI, CT, microscopy), covering a wide range of diverse objects of interest and clinical tasks. SamDSK relies on SAM to provide segmentation proposals. Therefore, SamDSK is especially applicable to those medical image segmentation tasks in which the regions of interest can be adequately addressed by segmentation proposals generated by SAM. It is important to note that our optimal matching formulation allows for the utilization of multiple segmentation proposals to cover one or more regions of interest. Consequently, SamDSK accommodates segmentation proposals that consist of over-segmented regions with respect to the regions of interest.

For illustration, we highlight several specific medical image segmentation tasks for which SamDSK is well-suited: (1) polyp segmentation in endoscopic images, (2) skin lesion segmentation in dermoscopic images, and (3) tumor region segmentation in ultrasound images. Fig.~\ref{fig:example_overview} presents visual examples of these tasks and the corresponding segmentation proposals generated by SAM. We conduct comprehensive experiments to demonstrate the effectiveness of SamDSK in tackling these segmentation tasks, as shown in the next section.

\section{Experiments}
We utilize three public datasets to demonstrate the effectiveness of our proposed SamDSK method. We compare SamDSK with several classical methods as well as state-of-the-art methods: (1) Pseudo-labels (PL)~\cite{lee2013pseudo}; (2) SP-Refine~\cite{thompson2022pseudo}: utilizing superpixels for refining pseudo-annotations; (3) Bidirectional Copy-Paste (BCP)~\cite{bai2023bidirectional}; (4) Uncertainty-aware Pseudo-label Selection (UPS)~\cite{rizve2020defense}: a multi-round SSL method based on uncertainty-aware pseudo-label selection. 

In addition to comparing with the known methods, we conduct ablation and additional studies to validate the effectiveness of the key components in SamDSK. In Section~\ref{sec:ab1}, we validate the effectiveness of utilizing SAM for sample selection. In Section~\ref{sec:ab2}, we validate the effectiveness of multi-round SSL with SAM. In Section~\ref{sec:ab3}, we demonstrate empirical coverage of the SAM-generated segmentation proposals with respect to ground truth annotations.

\subsection{Breast Cancer Segmentation in Ultrasound Images}
We first utilize the breast cancer segmentation task in ultrasound images~\cite{al2020dataset} to validate the effectiveness of our proposed SamDSK. This task is a binary segmentation problem of segmenting breast cancer regions in ultrasound images. We randomly split the original dataset into two equally-sized sets, one set for model training and the other set for testing. In the training set, we randomly select 30\% of samples and treat them as labeled images, and treat the remaining 70\% of samples as unlabeled images.\footnote{All the data split information will be published together with the code release.}
Two state-of-the-art medical image segmentation models, TransUNet~\cite{chen2021transunet} and HSNet~\cite{zhang2022hsnet}, are taken as the segmentation model $M$ for the experiments. We perform SamDSK for three rounds of processing (step-0 to step-4 as illustrated in Fig.~\ref{fig:overview} are one round of processing). For Eq.~(\ref{eq3}), both $v_1^{lower}$ and $v_1^{upper}$ are set as 1. In Table~\ref{BCS_results}, the results show that SamDSK outperforms the state-of-the-art methods for SSL. SamDSK yields considerably better segmentation performance than the closely related BPC method, indicating that SAM plays a critical role in sample selection and annotation construction of unlabeled images. In addition, the results show that the PL method may degrade the segmentation performance (see the TransUNet case) when the initial segmentation model incurs too many errors in pseudo-labels.

\subsection{Polyp Segmentation in Endoscopic Images}
Automatic polyp segmentation in endoscopic images can help improve the efficiency and accuracy of clinical screenings and tests for gastrointestinal diseases. Many deep learning (DL) based methods have been proposed for robust and automatic segmentation of polyps. Here, we utilize the SOTA polyp segmentation model HSNet~\cite{zhang2022hsnet} and the widely-used polyp segmentation model PraNet~\cite{fan2020pranet} for evaluating our proposed SamDSK method. HSNet uses the PVT backbone (Transformer-based) and PraNet uses the Res2Net backbone (CNN-based). We perform SamDSK for three rounds of processing. For Eq.~(\ref{eq3}), $v_1^{lower}$ is set as 1, 1, and 1 for the first, second, and third rounds, respectively, and $v_1^{upper}$ is set as 1, 2, and 3 for the first, second, and third rounds, respectively (the polyp class is set as class 1). That is, the constraints on the number of RoIs for the polyp class are gradually changed when we perform more rounds of model training and labeled set expansion, and we allow an increase in the number of segmentation proposals for the foreground-class segmentation. Following the data split settings in~\cite{fan2020pranet, zhou2023can}, 900 images from Kvasir~\cite{jha2020kvasir} and
550 images from CVC-ClinicDB~\cite{bernal2015wm} are randomly selected to form the training set. The remaining samples from these two datasets (i.e.,
Kvasir and CVC-ClinicDB) and the samples from
CVC-ColonDB~\cite{tajbakhsh2015automated}, ETIS~\cite{silva2014toward}, and CVC-300~\cite{vazquez2017benchmark} form the test set.

The results are reported in Table~\ref{polyp_results}, which show that SamDSK improves the segmentation performances of the HSNet and PraNet models for the settings where 10\% and 30\% of the total labeled samples from the training set are available. Compared to the classical pseudo-label method~\cite{lee2013pseudo} and the recently proposed methods, SamDSK achieves similar or better segmentation performances. Notably, we observe that SamDSK is stable and reliable in providing segmentation improvement, while some of the known methods can encounter situations where worse segmentation performances might be yielded in some cases. 



\subsection{Skin Lesion Segmentation in Dermoscopic Images}

The ISIC 2018 skin lesion segmentation dataset~\cite{codella2019skin} 
contains 2594 training dermoscopic images and 1000 test images for melanoma segmentation. Once again, we employ HSNet and PraNet for the experiments. We are aware that the image-level DSK for skin lesion segmentation in most of these images is that they contain only one region of interest. Consequently, we set $v_1^{lower}$ to 1 and $v_1^{upper}$ to 1 in Eq.~(\ref{eq3}) (the melanoma class is designated as class 1) throughout all the rounds of processing. The results in Table~\ref{isic_results} demonstrate that SamDSK enhances the segmentation performances of the state-of-the-art models when utilizing 10\% of the total labeled samples. SamDSK achieves a higher performance gain in comparison to the competing methods.





\subsection{Ablation and Additional Studies}\label{sec:ab}

\subsubsection{Effectiveness of Using SAM}\label{sec:ab1}
A key question that we seek to answer in this section is whether using the $\beta$ score obtained from optimal matching actually helps in selecting samples of better segmentation quality. Furthermore, we examine whether using the model's own segmentation map uncertainty and confidence serves as a better criterion for sample selection. In Fig.~\ref{fig:sample2}, we show that by using the $\beta$ score obtained from optimal matching, we can select samples with significantly better segmentation quality (higher mIoU) than using the model's own segmentation map uncertainty and confidence measures. For polyp segmentation, samples with the top 10\% of $\beta$ scores achieve a 95\% mIoU. For skin lesion segmentation, samples with the top 10\% of $\beta$ scores achieve a 92\% mIoU. Additionally, in Fig.~\ref{fig:points}, we provide visualization plots for ground truth annotations and segmentation predictions for samples in Case-1 and Case-2, which are identified after the optimal matching process. We observe that the samples in Case-1 (selected and added to the labeled set) exhibit a tighter alignment with their corresponding ground truth annotations than the samples in Case-2 (not yet added to the labeled set).

\subsubsection{Effectiveness of Multi-round SSL}\label{sec:ab2}
In Fig.~\ref{fig:SamDSK_plot1}, we show the segmentation performances with respect to the rounds of processing in SamDSK, along with that of the initial model (the baseline model). It is evident that the model's performance is improved through multiple rounds of processing, as more samples are added to the labeled set for model training/retraining.

\subsubsection{Performances of SAM in Generating Segmentation Proposals}\label{sec:ab3}
For the polyp segmentation task, we apply optimal matching between the segmentation proposals (generated by SAM) and the ground truth annotations. We use those segmentation proposals identified by optimal matching to construct segmentation annotation maps for each selected unlabeled sample. By comparing the generated annotation maps with the ground truths, we find that the segmentation annotations achieve an 89.1\% Dice coefficient. 
We apply the same procedure to the breast cancer segmentation dataset, and find that the Dice coefficient is 75.3\% for the benign cases and 64.4\% for the malignant cases. These results suggest that SAM performs better on endoscopic images than on ultrasound images (note that ultrasound images in the breast cancer segmentation dataset are in grayscale). This is probably due to the fact that SAM was trained using natural scene color images, which in appearance are closer to endoscopic images than ultrasound images. Additionally, in the case of ultrasound images, the benign cases consist of objects with more regular shapes and better contrast around the object boundaries, which may lead SAM to yield better coverage with its generated segmentation proposals.

\section{Conclusions}
In this paper, we proposed a new semi-supervised learning method, SamDSK, which utilizes the Segment Anything Model (SAM) for training a medical image segmentation model. SAM is employed in our proposed optimal matching system for selecting and refining segmentation predictions of unlabeled images (obtained from the current trained segmentation model). A multi-round iterative procedure of model training and labeled set expansion is performed on top of the optimal matching system to gradually improve the target segmentation model while enlarging the labeled training set. Experiments on three datasets, compared with several recently developed SSL methods, demonstrated the effectiveness and advantages of our proposed SamDSK method. Future work may study further expanding the applicable tasks of SamDSK to other medical imaging modalities and segmentation scenarios. 
\bibliography{sample}{}
\bibliographystyle{plain}

\end{document}